\def\year{2021}\relax
\documentclass[letterpaper]{article} 
\usepackage{aaai21}  
\usepackage{times}  
\usepackage{helvet} 
\usepackage{courier}
\usepackage[hyphens]{url}  
\usepackage{graphicx} 
\usepackage{natbib}  
\usepackage{caption} 
\usepackage{multirow} 
\usepackage{subcaption} 
\usepackage{amsfonts} 
\usepackage{amsmath} 
\title{Detecting Concept Drift in the Presence of Sparsity - A Case Study of Automated Change Risk Assessment System}

\iftrue
\author{
    Vishwas Choudhary,\textsuperscript{\rm 1}\thanks{Work done during internship at Walmart Global Tech, Bangalore, India.}
    Binay Gupta, \textsuperscript{\rm 2}
    Anirban Chatterjee, \textsuperscript{\rm 2}
    Subhadip Paul, \textsuperscript{\rm 2}
    Kunal Banerjee, \textsuperscript{\rm 2}
    Vijay Agneeswaran \textsuperscript{\rm 2} \\
}
\affiliations{
    \textsuperscript{\rm 1} Indian Institute of Technology, Kanpur, India\\
    \textsuperscript{\rm 2} Walmart Global Tech, Bangalore, India\\
    
    vishwasc@iitk.ac.in, 
    \{binay.gupta, anirban.chatterjee, subhadip.paul0, kunal.banerjee1, vijay.agneeswaran\}@walmart.com
}
\fi

\begin{document}

\maketitle

\begin{abstract}
\begin{quote}
Missing values, widely called as \textit{sparsity} in literature, is a common characteristic of many real-world datasets.
Many imputation methods have been proposed to address this problem of data incompleteness or sparsity.
However, the accuracy of a data imputation method for a given feature or a set of features in a dataset is highly dependent on the distribution of the feature values and its correlation with other features.
Another problem that plagues industry deployments of machine learning (ML) solutions is concept drift detection, which becomes more challenging in the presence of missing values.
Although data imputation and concept drift detection have been studied extensively, little work has attempted a combined study of the two phenomena, i.e., concept drift detection in the presence of sparsity.
In this work, we carry out a systematic study of the following:
(i) different patterns of missing values,
(ii) various statistical and ML based data imputation methods for different kinds of sparsity,
(iii) several concept drift detection methods,
(iv) practical analysis of the various drift detection metrics,
(v) selecting the best concept drift detector given a dataset with missing values based on the different metrics.
We first analyze it on synthetic data and publicly available datasets, and finally extend the findings to our deployed solution of automated change risk assessment system.
One of the major findings from our empirical study is the absence of supremacy of any one concept drift detection method across all the relevant metrics.
Therefore, we adopt a majority voting based ensemble of concept drift detectors for abrupt and gradual concept drifts.
Our experiments show optimal or near optimal performance can be achieved for this ensemble method across all the metrics.
\end{quote}
\end{abstract}

\section{Introduction}\label{sec:intro}
There are many reasons why missing values often occur in a dataset -- it may be because of erroneous data entry process, irregular data collection or intentionally not supplied by the users, especially when they are filling forms that have non-mandatory fields.
Missing values can make the downstream tasks, such as, predicting the value of a target variable, much more challenging.
Since performance of all algorithms degrade in the presence of sparsity, and many a times algorithms are not even designed to handle missing data, one may either ignore rows that have missing values altogether -- however, this may cause problems, especially when the sparsity level is high, or use some data imputation technique before applying the algorithm.
The situation is further aggravated when one tries to detect concept drifts on such datasets with sparsity -- we experienced this firsthand when we tried to do the same with our deployment of machine learning solution for \textit{change risk assessment}~\cite{CD_Walmart}.
This solution is currently operational for changes targeted across Walmart's US, UK and Mexico stores, US Sam's Clubs and e-Commerce.
Post deployment, the production team has confirmed that the number of major incidents has been reduced by 33\% with net savings ranging into multi-million dollars as in Q2 of 2021.
Note that there may be other factors, e.g., software design changes, that have contributed to the savings; however, it is acknowledged that our ML based prediction system has been the primary contributor.
To improve the performance further, we studied concept drift detection in the presence of sparsity, and based on our findings, the main contributions of this work are as follows:
\begin{itemize}
 \item Provide an empirical guideline on applying data imputation given a data distribution and a sparsity pattern.
 \item Suggest various metrics for concept drift detection -- we found that some of the metrics in literature can be misleading in practice, and propose some new ones that we found helpful.
 \item Provide a majority voting based ensemble of concept drift detectors for abrupt and gradual drifts that perform well across the whole spectrum of concept drift detection metrics -- a feat that individual concept drift detectors may not attain.
\end{itemize}

\section{Preliminaries}\label{sec:prelim}

\subsection{Types of Missingness}
The nomenclature for the types of missing values was introduced by Rubin in~\cite{Rubin} that is considered a defacto standard in any kind of statistical analysis with incomplete data.
This nomenclature distinguishes between three cases: 
\begin{itemize}
 \item \textbf{Missing Completely At Random (MCAR).} In MCAR, the missingness is completely independent of the data.
 \item \textbf{Missing At Random (MAR).} In MAR, the probability of missingness depends only on observed values.
 \item \textbf{Missing Not At Random (MNAR).} In MNAR, the probability of missingness  depends on the unobserved values, and therefore it leads to important biases in the data.
\end{itemize}

\subsection{Data Imputation Techniques}\label{sec:data_impute}
These techniques can be broadly differentiated into two categories.
First, those techniques that look into a single feature at a time -- \textbf{mean}, \textbf{median}, \textbf{mode} and \textbf{zero} -- these replace the missing values with the mean, median, highest occurring (mode) and a constant value zero, respectively.
Second, those techniques that look into other features as well for correlation -- k-nearest neighbours (\textbf{kNN}), \textbf{iterative imputer}~\cite{iterativeImputer}, \textbf{soft impute}~\cite{softImpute} and \textbf{optimal transport}~\cite{optimalTransport}.

\subsection{Concept Drift}
Concept drift is the change in the joint probability
distribution for input $X$ and label $y$ between two time points $t_0$ and $t_1$ \cite{CD_Walmart}.

\subsection{Concept Drift Detection Algorithms}\label{sec:algo}
The concept drift detection algorithms discussed here are as follows.
Page-Hinkley (\textbf{PH})~\cite{pageHinkley} signals concept drift when the difference of the observed values from the mean crosses over a user-defined threshold.
Drift Detection Method (\textbf{DDM})~\cite{ddm} detects concept drifts in streams by analyzing the error rates and their standard deviation; if the error rate increases, then DDM concludes that the current predictor is outdated.
Early Drift Detection Method (\textbf{EDDM}) monitors the distance between two consecutive errors, rather than the error rate.
Consequently, when the concepts remain stationary the distance grows larger, whereas, a decrease in the distance signals drift.
Heoffding’s inequality based Drift Detection Method (HDDM)~\cite{hddm} has two variants: \textbf{HDDM\textsubscript{A}} that uses moving averages to detect drifts, and \textbf{HDDM\textsubscript{W}} that uses exponentially weighted moving averages to detect drifts.
ADaptive WINdowing (\textbf{ADWIN})~\cite{adwin} maintains two sub-windows, one for historic data and the other for new data; a significant difference between the means of these sub-windows indicates a concept drift.
Kolmogorov-Smirnov WINdowing (\textbf{KSWIN})~\cite{kswin} maintains a sliding window of fixed size $n$ where the last $r$ samples represent the latest concept.
A concept drift is detected if Kolmogorov-Smirnov test between the two distributions of $r$ and $n-r$ samples yields a significant difference.

\subsection{Metrics for Concept Drift}\label{sec:metrics}
It is important to note that the first two metrics apply to the model that makes the predictions, while the latter four apply to the concept drift detector (CDD).
Note that the last two metrics have been proposed by us.

\noindent \textbf{Prequential Error:} For a given sequence of instances $1,2,\ldots$, it is the accumulated sum, i.e. $e_i = \frac{1}{i}\sum_{k=1}^{i}L(y_k,\hat{y}_k)$, of a loss function $L$ between the prediction $\hat{y}_k$ and the  observed value $y_k$.

\noindent \textbf{Accuracy:} It is the fraction of the accurate predictions made to the total number of predictions.

\noindent \textbf{Average Detection Delay (ADD):} It is average of the distances (in terms of instances) between the actual and the predicted concept drifts.

\noindent \textbf{True Positive Rate (TPR):} Fraction of detected drifts which were true, i.e., within acceptable detection interval (ADI), to the total number of drifts detected.\\
Due to detection delays, in our experiments, we keep the ADI four times the drift width as per existing literature~\cite{fastHoeffdig,accurate}.

\noindent \textbf{True Positives per Drift (TPD):} Fraction of detected drifts which were true to the total number of actual drifts.\\
It is possible that a CDD declares $2$ or more drifts within the ADI, and accordingly, its TPR may be high which can be misleading; therefore, we introduce TPD whose optimal value is $1$ (though it can be higher or lower), i.e., CDD declares a drift exactly once for every actual drift in the dataset.

\noindent \textbf{Drift Count:} Total number of actual drifts detected.\\
Since detecting multiple drifts within the ADI for an actual drift may offset for the case when no drift is detected for another actual drift, we included drift count in our set of metrics.

\section{Background \& Motivation}\label{sec:background}
Concept drift problem exists in many real-world situations such as automated change risk assessment in technology driven industry. This can result in poor and degrading predictive performance in predictive models that assume a static relationship between input and output variables. 
\par One brute-force way to mitigate this problem is to periodically train the static model with more recent data. 
However, the cost factor, such as \textit{Computational Cost}, \textit{Labor Cost} and \textit{Implementation Cost},  puts up a significant impediment towards retraining model frequently.
\par A more elegant way to approach this problem is to algorithmically detect concept drift in data and retrain the model depending on the outcome of the algorithm.
There are different categories of drift detection algorithms such as error rate-based drift detection algorithms, distribution-based drift detection algorithms, and few others~\cite{cdrift-review}. 
However, such drift detection methods suffer from severe impairment in presence of data sparsity. 
Non-consideration of inaccuracies in the imputation methods in estimating the underlying data distribution is at the core of this problem~\cite{cdrift}.
A fuzzy distance estimation based method for detecting concept drifts in the presence of missing values has been addresses in~\cite{cdrift}.
To formally discuss the impact of data sparsity on drift-detection methods, we need to delve deep into the following questions:
\begin{itemize}
    \item \textbf{Question 1.} How can imputing missing values perturb the original distribution of the data?
    \item \textbf{Question 2.} How can a drift detection method be impacted by the perturbed distribution of data?
\end{itemize}
To analyze \textbf{Question 1}, we first introduce a few notations.
We assume that we are given a sample space $\chi$, a finite set of data points.
From this sample space, we are given a set of positive labeled points $\{x_1, \cdot \cdot \cdot , x_m\}$.
We assume these labeled points are drawn i.i.d. from some unknown target distribution $\pi$ over $\chi$.
We assume that $f(x) = \{f_1(x), \cdot \cdot \cdot , f_n(x)\}$ represents a data-point $x$ which is characterized by a set of features $\{f_j : \chi \xrightarrow{} \mathbb{R}\}$ for $j \in \{1, \cdot \cdot \cdot , n\}$.
We define the missing indicator variable $m_j(x)$ as follows:
\[   
m_j(x)=
     \begin{cases}
       \ $1$ &\quad\text{if $f_j(x)$ is observed}\\
       \ $0$ &\quad\text{if $f_j(x)$ is missing}\\
     \end{cases}
\]
In any setting of missingness, such as, MCAR, MAR or MNAR, the true expectation of the data can be quite deviant from the observable expectation depending on the accuracy of missing value imputation. Having $p(m_j(x))$ to represent the prior on $m_j(x)$ and $p(m_j (x) \lvert f(x))$ to represent the conditional distribution of missingness which can be MCAR, MAR or MNAR, we can write applying Bayes rule:
\begin{equation}
\begin{aligned}
\displaystyle p(f(x) \lvert m_j (x)) = \frac{p(m_j (x) \lvert f(x)) p (f(x))}{p(m_j(x))}
\end{aligned}
\end{equation}
We can estimate the empirical expected value of a single feature $f_j$ the following way:
\begin{equation}
\begin{aligned}
\displaystyle \hat{E_1}(f_j) &= \frac{\sum_{\substack{x \in \chi}}p(f(x)) f_j(x)}{\sum_{x \in \chi}p(f(x)) m_j(x)} \\
&= \frac{\sum_{x \in \chi}p(f(x)|m_j (x))f_j (x)}{\sum_{x \in \chi}p(f(x)) m_j(x)}\\
&= \sum_{x \in \chi}p(f(x)|m_j (x) = 1)f_j(x)\\
&= E_{p(f(x)|m_j(x)=1)}[f_j ]
\end{aligned}
\end{equation}
In other words, the estimated expected value of a feature is with respect to the conditional probability $p(f(x) \lvert m_j (x))$ instead of $p(f(x))$. We now discuss the situation when we impute the missing values. Let $I_j(x \lvert \chi,(f_1, m_1), . . . ,(f_n, m_n))$, which we denote as $I_j(x)$ for simplicity, represent any imputation method invoked when $f_j (x)$ is missing. In other words, we use $I_j(x)$ to approximate the ground truth of $f_j(x)$ when it is missing.
\begin{equation}
\begin{aligned}
\displaystyle \hat{E_2}(f_j) &= \sum_{\substack{x \in \chi \\ m_j(x) = 1}}p(f(x))f_j(x) + \sum_{\substack{x \in \chi \\ m_j(x) = 0}}p(f(x))I_j(x)\\
&= w_1 E_{p(f(x)|m_j(x)=1)}[f_j] + w_2 E_{p(f(x)|m_j(x)=0)}[I_j]
\end{aligned}
\end{equation}
where
\begin{equation}
\begin{aligned}
\displaystyle w_1 &= \sum_{x \in \chi}p(f(x)) m_j(x)\\
w_2 &= \sum_{x \in \chi}p(f(x)) (1 - m_j(x))
\end{aligned}
\end{equation}
Observe that two weighted expectations are contributing to the estimated expectation of feature $f_j$. The first term in equation 3 represents the expectation of feature $f_j$ estimated on the basis of only the observed values. The second term, however, is the expectation over another conditional distribution of some imputed feature $I_j$. The deviation between the imputed feature value $I_j$, and the corresponding ground truth  depends on the accuracy of the imputation method. Notice also that higher is the degree of sparsity, more pronounced is the effect of $I_j$ on $\hat{E_2}(f_j)$.
\par Equation 3 essentially means 
that,
in effect of missing value imputation, $p(f_j)$, the marginal distribution of every feature, $p(f(x))$, the overall data distribution and $p(y \lvert f(x))$, the conditional distribution where $y$ represents the label of the data-point represented by $f(x)$ admit some degree of perturbation. 
\par For \textbf{Question 2}, we first take an example of an error-rate based method, such as, DDM~\cite{ddm}. The essence of this method lies in comparing the error rate of a classifier between two consecutive batches of data-points of length $W$ and checking if the change in error-rate of the classifier between these two batches is statistically significant.
\par Suppose a sequence of examples, in the form of pairs $(x_i, y_i)$. For each example, the actual classification model predicts $\hat{y}_i$, that can be True or False. For a set of examples, the error is a random variable from Bernoulli trials~\cite{Forbes}. The Binomial distribution gives the general form of the probability for the random variable that represents the number of errors in a sample of $n$ examples. For each point $i$ in the sequence, the error-rate is the probability of observed False, $p_i$, with standard deviation given by $s_i = \sqrt{p_i(1 - p_i)/i}$.
Statistical theory~\cite{Mitchell} guarantees the decline in error rate of the learning algorithm ($p_i$) with the increase of $i$ on the condition of stationarity of class distribution. Suppose the length of the first batch is $n$ and that of the second batch is $m$ with proportion of error being $\theta_X$ and $\theta_Y$, respectively. $X$ and $Y$ represent two random variables satisfying $X \sim Binomial (n, \theta_X)$ and $Y \sim Binomial (m, \theta_Y)$. Error-rate based concept drift detection methods, such as,  DDM, end up constructing a test to validate the following null hypothesis:
\begin{equation}
\begin{aligned}
\displaystyle H_0 : \theta_X = \theta_Y
\end{aligned}
\end{equation}
\par The limitation of the above approach becomes prominent in the presence of a high degree of sparsity in the two consecutive sequences of data samples. As explained in previous section, the distributions of the data of those two batches get perturbed following missing value imputation. 
Therefore, the binomial distribution which is used to model the error-rate in those two batches also ends up being perturbed and eventually results in wrong perception of concept drift.
\par Data distribution-based drift detection algorithms are considered the second largest category of drift detection methods where a distance function is employed to quantify the dissimilarity between the distributions of two consecutive sequences or batches of data samples. As the missing value imputation perturbs the batches of data samples, this category of algorithms suffers from similar problem as mentioned above in the presence of high degree of sparsity.

\par Therefore, it warrants us to develop a strategy to ensure a guarantee, at least to some extent, on the performance of the drift detection method even in the presence of various non-ideal data characteristics, such as, high degree of sparsity. 

\section{Methodology}\label{sec:method}
Our methodology involves the following broad steps.

\begin{table}[tbh]
\centering
\begin{tabular}{c|c|c|c}
{\bf Distribution}  & {\bf MCAR} & {\bf MAR}  & {\bf MNAR}\\
\hline
Normal      & Mean   & Mean   & Mean  \\
\hline
Uniform     & Mean   & Mean   & Mean  \\
\hline
Chi-squared & Mean   & Mean   & Mean  \\
\hline
Cauchy      & Median & Median & Median\\
\hline
Binomial    & Median & Median & Median\\
\hline
Multi-variate & kNN50 ($<$30\%)  & kNN100 & kNN50\\
normal        & kNN100 ($>$30\%) &        & 
\end{tabular}
\caption{Distribution-wise best data imputation technique.}
\label{tab:dist}
\end{table}
\noindent \textbf{1. Find distribution-wise best data imputation scheme.}
We started by creating various distributions (one may think of these as synthetic datasets with a single feature except for the multi-variate normal case), and applied MCAR, MAR and MNAR sparsities at different levels -- $5\%, 10\%, 20\%, \ldots, 60\%$.
The data imputation techniques applied were mean, median, mode, zero (constant), and kNN (we tried multiple values of k).
Our findings are summarized in Table~\ref{tab:dist}.
Note that for some distributions, some of the data imputation schemes may be identical, e.g., mean, median and mode for normal distribution; however, we mention only one of these identical schemes for a given distribution in this table.
Surprisingly, other than the multi-variate normal distribution, the best performing scheme is identical irrespective of the type of missingness and the sparsity level.

\begin{figure*}[tbh]
\centering
\includegraphics[width=0.9\linewidth]{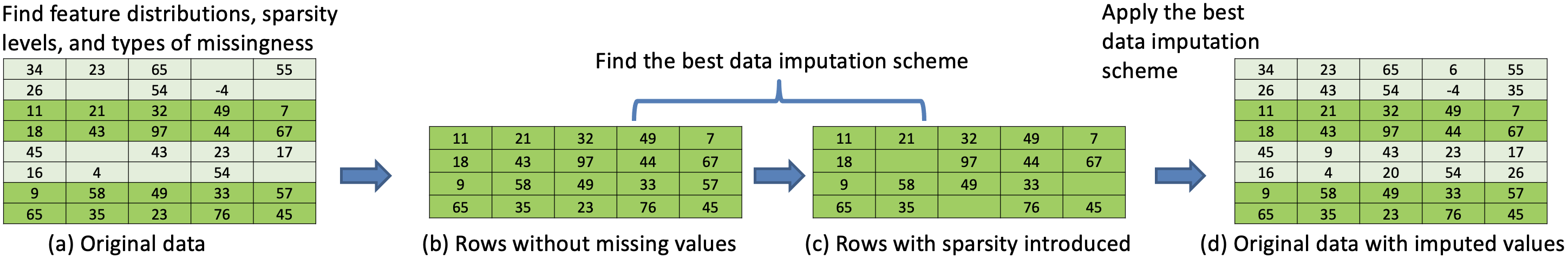}
\caption{Finding the best data imputation scheme for a given dataset with sparsity.}
\label{fig:data_imputation}
\end{figure*}
\noindent \textbf{2. Find the best data imputation scheme for a given dataset.}
For a given dataset with sparsity, we first identify what is the level of sparsity for each feature.
Next we replace each missing value with 0 and non-missing value with 1 and apply runs test for randomness~\cite{runsTest,multiRunsTest}.
Given our application domain and the corresponding data, which is mostly obtained through forms filled in by human change requestors who tend to leave out non-mandatory fields and thus leading to sparsity, we ruled out the possibility of MCAR (although we do perform experiments with MCAR on synthetic datasets in subsequent section), and strongly believe that there is a (possibly hidden) pattern to the missingness in our data.
We envision that most domain experts should be able to use prior knowledge to tell whether MCAR is indeed a possibility for their use cases or not -- in our experience, it is a rare phenomenon.
Therefore, if the test declares that the data is missing at random, then we conclude that it is a case of MNAR, i.e., we are currently not collecting the variables on which the missingness depend; otherwise, we consider it a case of MAR.
We also use quantile-quantile (Q-Q) plots to find the probability distribution of each feature.
Subsequently, we isolate the rows that do not have any missing values.
Having already identified the sparsity level and the type of missingness, we apply the same on these rows (without sparsity).
Since we know the original values for these newly introduced missing values, we use the same to find the best data imputation scheme among the winner identified in the previous step and correlation-dependent imputation techniques mentioned in Section~\ref{sec:data_impute}.
We use root-mean-square error (RMSE) to find the best scheme.
This strategy is captured in Figure~\ref{fig:data_imputation}.

\noindent \textbf{3. Apply the majority voting based ensemble of concept drift detectors on the dataset with imputed values.}
The central idea behind our approach is to construct an ensemble of multiple CDDs and infer on concept drift based on the majority voting among the individual CDDs. The intuition is that the majority voting strategy will provide a lower bound on the resulting accuracy by taking advantage of the strengths of individual CDDs in diverse situations.
\par On a formal note, consider a concept drift detection problem in which an algorithm $CD$ takes two consecutive sequences of data instances, $S_1 = \{\mathbf{x}_i\}_{i=1}^n$ and $S_2 = \{\mathbf{x}_j\}_{j=1}^m$, from a data stream and predicts if drift exists between $S1$ and $S_2$. We have $\hat{y}_{CD} \in \{-1, +1\}$ to represent the prediction of the drift detection algorithm $CD$ where -1 and +1 represent absence and presence of drift respectively. The overall decision is based on the average prediction of the base CDDs. Let the average prediction of the base CDDs, $CD_p$, be the score $\phi \in [-1, +1]$:
\begin{equation}
\begin{aligned}
\displaystyle \phi(S_1,S_2) = \frac{1}{N} \sum_{p=1}^N CD_p (S_1,S_2)
\end{aligned}
\end{equation}
In the usual case, if the score is negative then the overall decision is
$\hat{y}_{CD} = -1$, and if positive then $\hat{y}_{CD} = +1$, i.e. $\hat{y}_{CD} (S_1, S_2) = sign(\phi(S_1,S_2))$.
\par To analyze the properties of the ensemble CDD, consider a generic framework in which the ensemble CDD, $\phi$, takes the overall decision:
\[   
\hat{y}_{CD} (S_1, S_2)=
     \begin{cases}
       \ $-1$ &\quad\text{if $\phi(S_1,S_2) \leq -t$}\\
      \ \text{reject} &\quad\text{if $-t < \phi(S_1,S_2) < t$}\\
       \ $+1$ &\quad\text{if $\phi(S_1,S_2) \geq t$}
     \end{cases}
\]
where $t \geq 0$ is a rejection threshold and $[-t, t]$ is the most uncertain region for the ensemble drift detector $\phi$. Define a random variable $z = y\phi \in [-1, +1]$ where $y$ represents the ground truth of the existence of drift between $S_1$.and $S_2$ and admits value either $+1$ or $-1$ depending on the existence of drift while $\phi$ is the output of the ensemble drift detector as defined above. Due to the encoding $\hat{y}_{CD} \in \{-1, +1\}$, $z$ is negative for incorrect
predictions and positive for correct predictions. If $z$ is in the range $[-t, +t]$, it causes the ensemble drift detector to suffer from indecision. With $z$ being a random variable, the probability of wrong prediction
\begin{equation}
\begin{aligned}
\displaystyle P_E &= Pr [ z \leq -t]\\
\end{aligned}
\end{equation}
and the probability of rejection
\begin{equation}
\begin{aligned}
\displaystyle P_R &= Pr [ - t < z < t]\\
\end{aligned}
\end{equation}
We now define the risk of the ensemble drift detector $\phi$ as below:
\begin{equation}
\begin{aligned}
\displaystyle R_\phi &= c_1 P_E + c_2 P_R \\ 
&= c_1 Pr [ z \leq -t] + c_2 Pr [ - t < z < t]\\
&= (c_1 - c_2) Pr [ z \leq -t] + c_2 Pr [z < t]
\end{aligned}
\end{equation}
Here $c_1$ and $c_2$ represent the cost of wrong prediction and rejection or indecision respectively. We now look to find an upper bound of $R_\phi$ to analyze the worst case situation. As discussed in~\cite{Breiman}, we have
\begin{equation}
\begin{aligned}
\displaystyle var(z) \leq \Bar{\rho}(1 - \mu_z^2)
\end{aligned}
\end{equation}
where $\Bar{\rho}$ represents the average pairwise correlation between base CDDs and $\mu_z = E[z]$. Following Cantelli's inequality (Wu \textit{et al.}, 2021), we have
\begin{equation}
\begin{aligned}
\displaystyle Pr [ z - \mu_z \leq -a ] \leq \frac{1}{1 + \frac{a^2}{var(z)}}, \ \ a > 0
\end{aligned}
\end{equation}
Setting $ a = \mu_z + t$, we obtain
\begin{equation}
\begin{aligned}
\displaystyle Pr [ z \leq -t ] 
&\leq \frac{1}{1 + \frac{(\mu_z + t)^2}{\Bar{\rho}(1 - \mu_z^2)}}, \ \ \mu_z > -t
\end{aligned}
\end{equation}
Similarly, using the inequality 10, we have
\begin{equation}
\begin{aligned}
\displaystyle Pr [ z \leq t ] &\leq \frac{1}{1 + \frac{(\mu_z - t)^2}{\Bar{\rho}(1 - \mu_z^2)}}, \ \ \mu_z > t
\end{aligned}
\end{equation}
Thus combining 9, 12 and 13, we can find the upper bound of $R_\phi$ as below:
\begin{equation}
\begin{aligned}
\displaystyle R_\phi \leq \frac{c_1 - c_2}{1 + \frac{(\mu_z + t)^2}{\Bar{\rho}(1 - \mu_z^2)}} + \frac{c_2}{1 + \frac{(\mu_z - t)^2}{\Bar{\rho}(1 - \mu_z^2)}}, \ \ \mu_z > t
\end{aligned}
\end{equation}
In our model, we set $t = 0$ which gives,
\begin{equation}
\begin{aligned}
\displaystyle R_\phi \leq \frac{c_1}{1 + \frac{\mu_z^2}{\Bar{\rho}(1 - \mu_z^2)}}, \ \ \mu_z > 0
\end{aligned}
\end{equation}
 
Observe that the above inequality applies only when $\mu_z > 0$ and the upper bound of $R_\phi$ comes down with the decrease in $\Bar{\rho}$. In other words, with more diversity among the base CDDs, the ensemble, $\phi$ as defined in equation 6, becomes less prone to wrong prediction about concept drift even if the base CDDs suffer from deterioration in performance due to sparsity.

\section{Experimental Setup \& Results}\label{sec:result}
\subsection{Dataset Description}
\begin{figure*}[tbh]
 \begin{minipage}[t]{.37\textwidth}
  \centering
  \includegraphics[width=\textwidth]{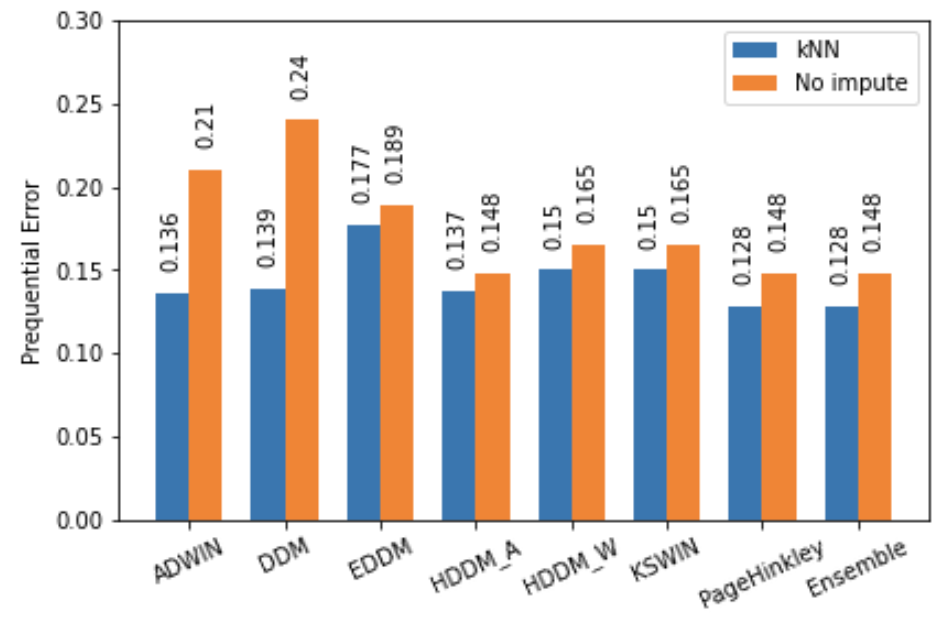}
  \subcaption{Prequential error.}
 \end{minipage}
 \hfill
 \begin{minipage}[t]{.37\textwidth}
  \centering
  \includegraphics[width=\textwidth]{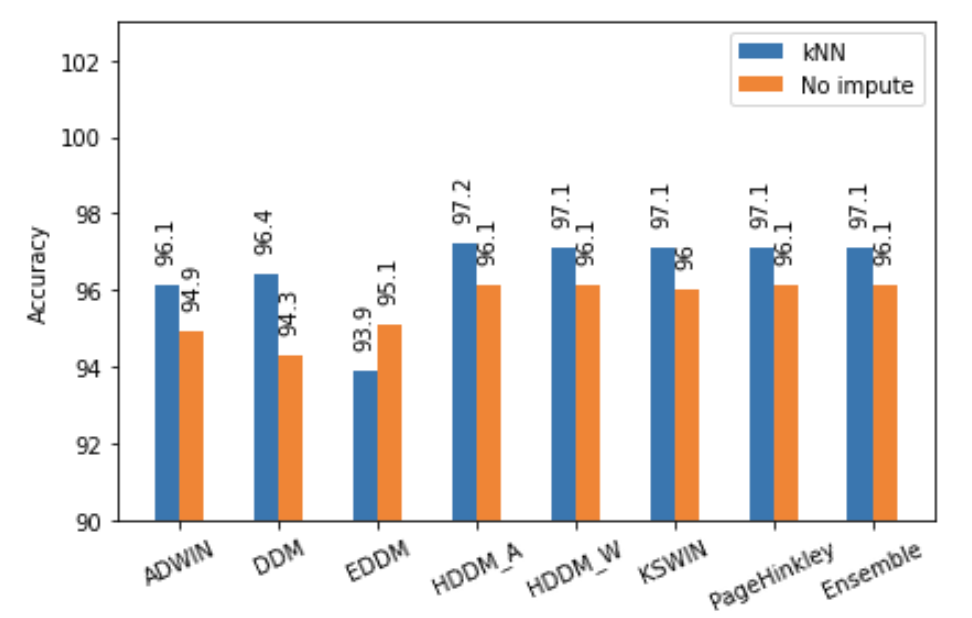}
  \subcaption{Accuracy.}
 \end{minipage}
 \newline
 \begin{minipage}[t]{.37\textwidth}
  \centering
  \includegraphics[width=\textwidth]{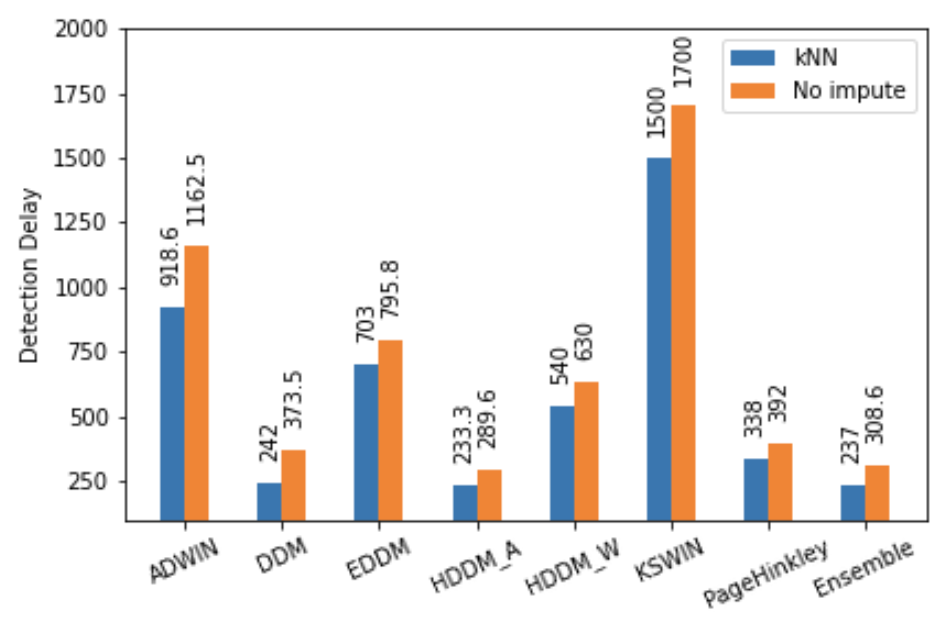}
  \subcaption{Average detection delay.}
 \end{minipage}
 \hfill
 \begin{minipage}[t]{.37\textwidth}
  \centering
  \includegraphics[width=\textwidth]{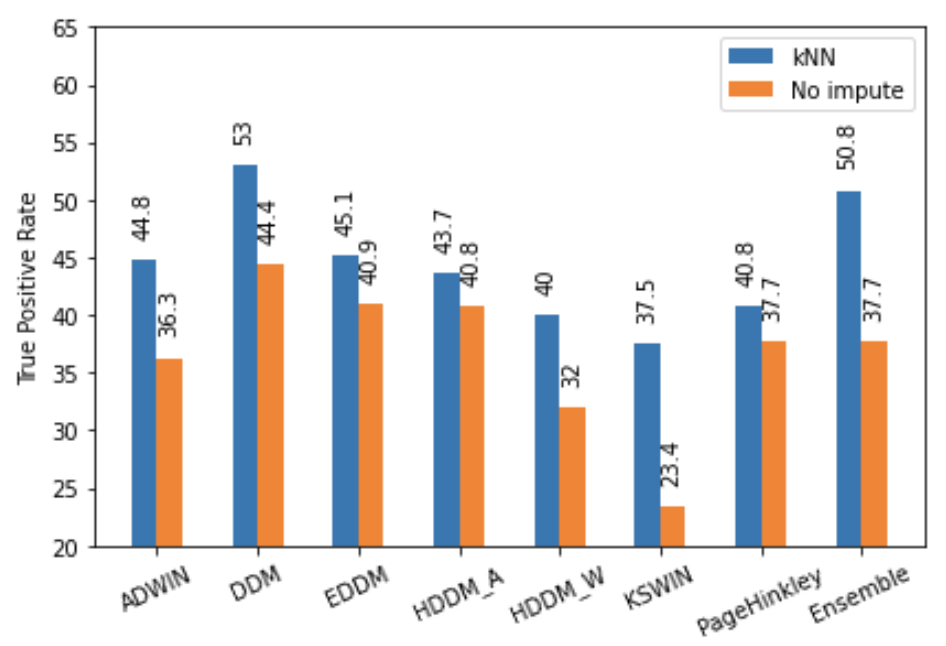}
  \subcaption{True positive rate.}
 \end{minipage}
 \newline
 \begin{minipage}[t]{.37\textwidth}
  \centering
  \includegraphics[width=\textwidth]{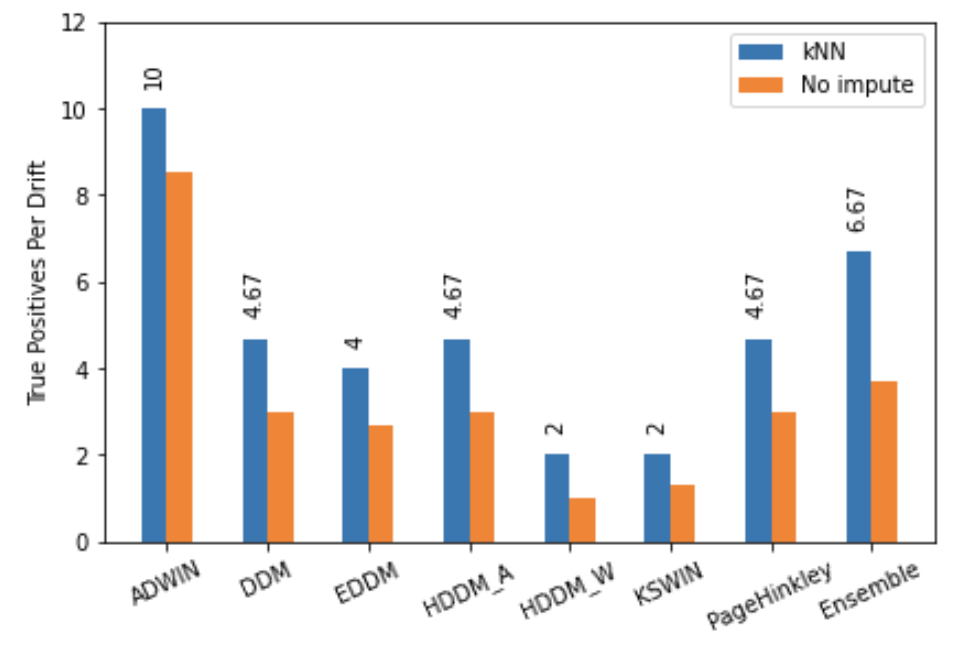}
  \subcaption{True positives per drift.}
 \end{minipage}
 \hfill
 \begin{minipage}[t]{.37\textwidth}
  \centering
  \includegraphics[width=\textwidth]{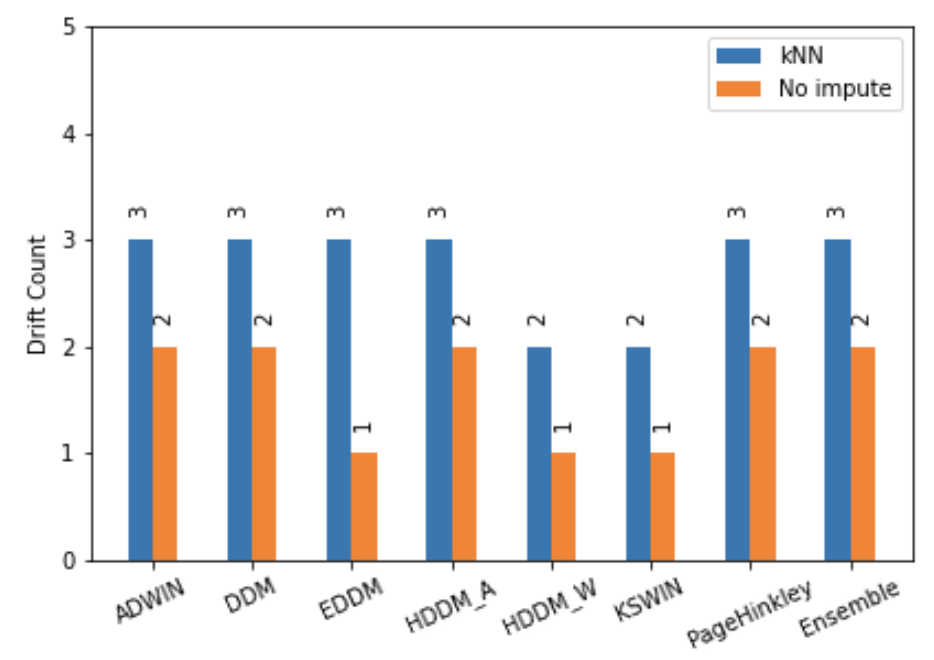}
  \subcaption{Drift count.}
 \end{minipage}
 \caption{Performance of CDDs with and without data imputation for change risk assessment data.}
 \label{fig:metrics}
\end{figure*}
We use the publicly available Harvard dataverse~\cite{harvardDataverse} that contains synthetic datasets for abrupt and gradual concept drift detection.
For our change risk assessment project, we have $\sim$50K samples (change requests) that are labelled as either ``risky'' (i.e., potentially may lead to some major incident) or ``not risky'' -- thus, our primary task is binary classification.
Now, we take the $\sim$50K samples and shuffle these to remove any pre-existing concept drifts in the data, and then create separate datasets of abrupt and gradual concept drifts by flipping the class labels; we tried various combinations of sample points where the drifts are introduced and various drift widths.
Note that this process of shuffling the original dataset (which ideally has no adverse effect on the original classifier), and then introducing drifts artificially is a common process for real-world datasets because in reality, it is not always feasible to clearly identify the onset and width of a concept drift~\cite{challengesRealData,reliableDetection}.

\subsection{Experimental Results}
Initially, we check whether data imputation aids in concept drift detection or not.
On experimenting with Harvard dataverse and our data, we find that data imputation had a positive effect on all the metrics mentioned in Section~\ref{sec:metrics} and for all the algorithms in Section~\ref{sec:algo} -- this is shown in Figure~\ref{fig:metrics} for our data, which had lowest RMSE for kNN4.

\begin{figure*}[tbh]
 \begin{minipage}[t]{.4\textwidth}
  \centering
  \includegraphics[width=\textwidth]{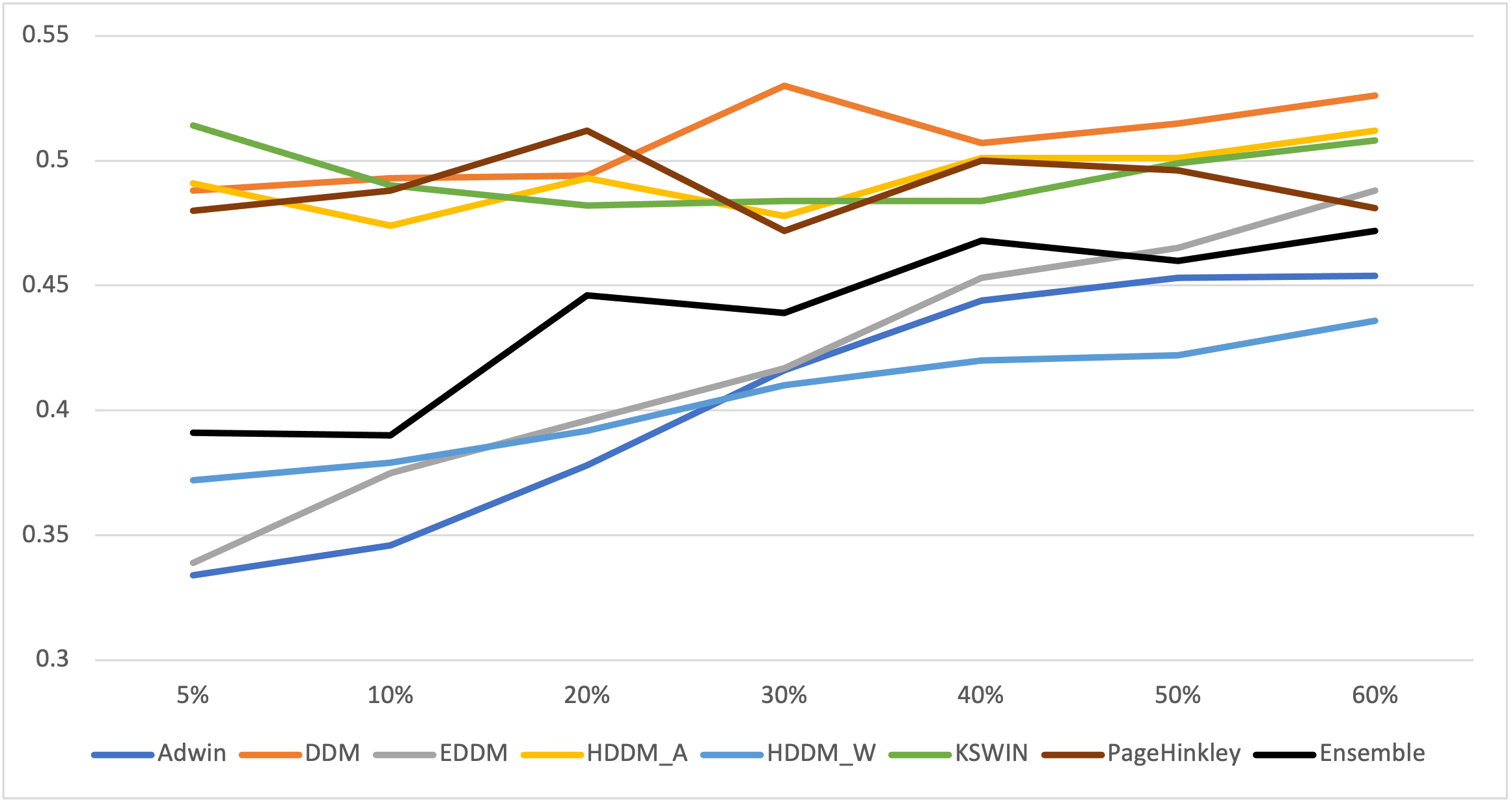}
  \subcaption{Prequential error.}
 \end{minipage}
 \hfill
 \begin{minipage}[t]{.4\textwidth}
  \centering
  \includegraphics[width=\textwidth]{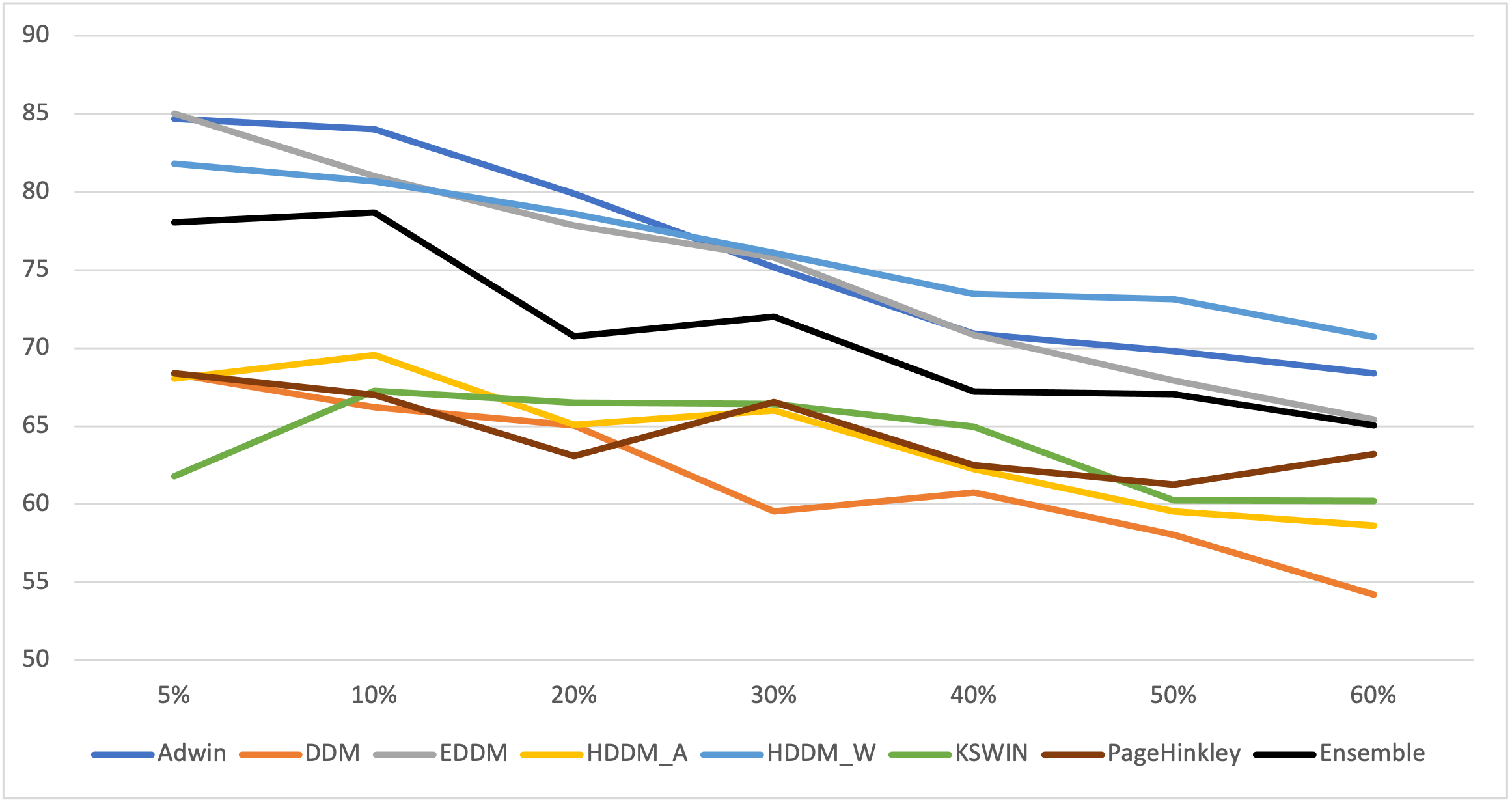}
  \subcaption{Accuracy.}
 \end{minipage}
 \newline
 \begin{minipage}[t]{.4\textwidth}
  \centering
  \includegraphics[width=\textwidth]{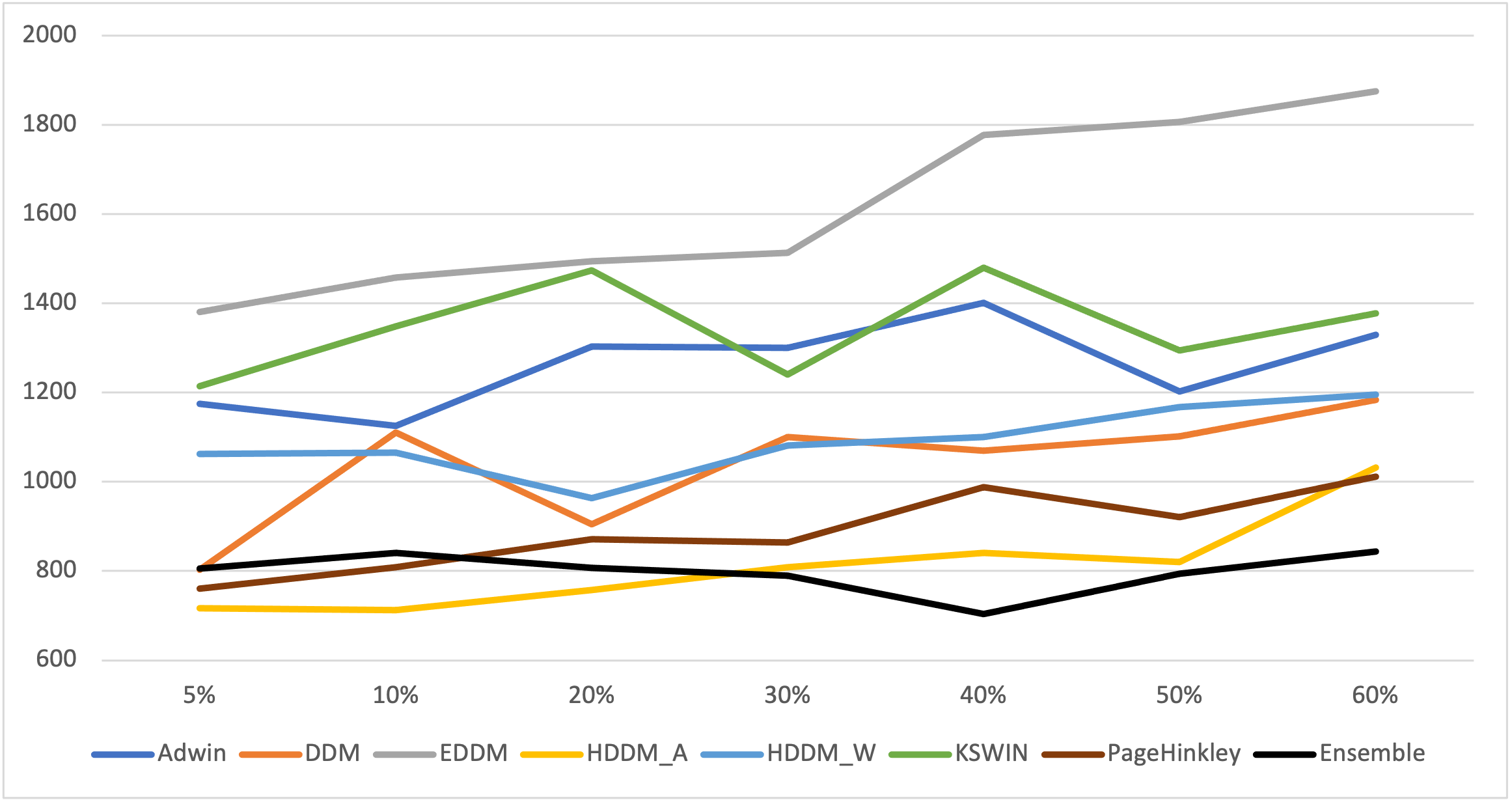}
  \subcaption{Average detection delay.}
 \end{minipage}
 \hfill
 \begin{minipage}[t]{.4\textwidth}
  \centering
  \includegraphics[width=\textwidth]{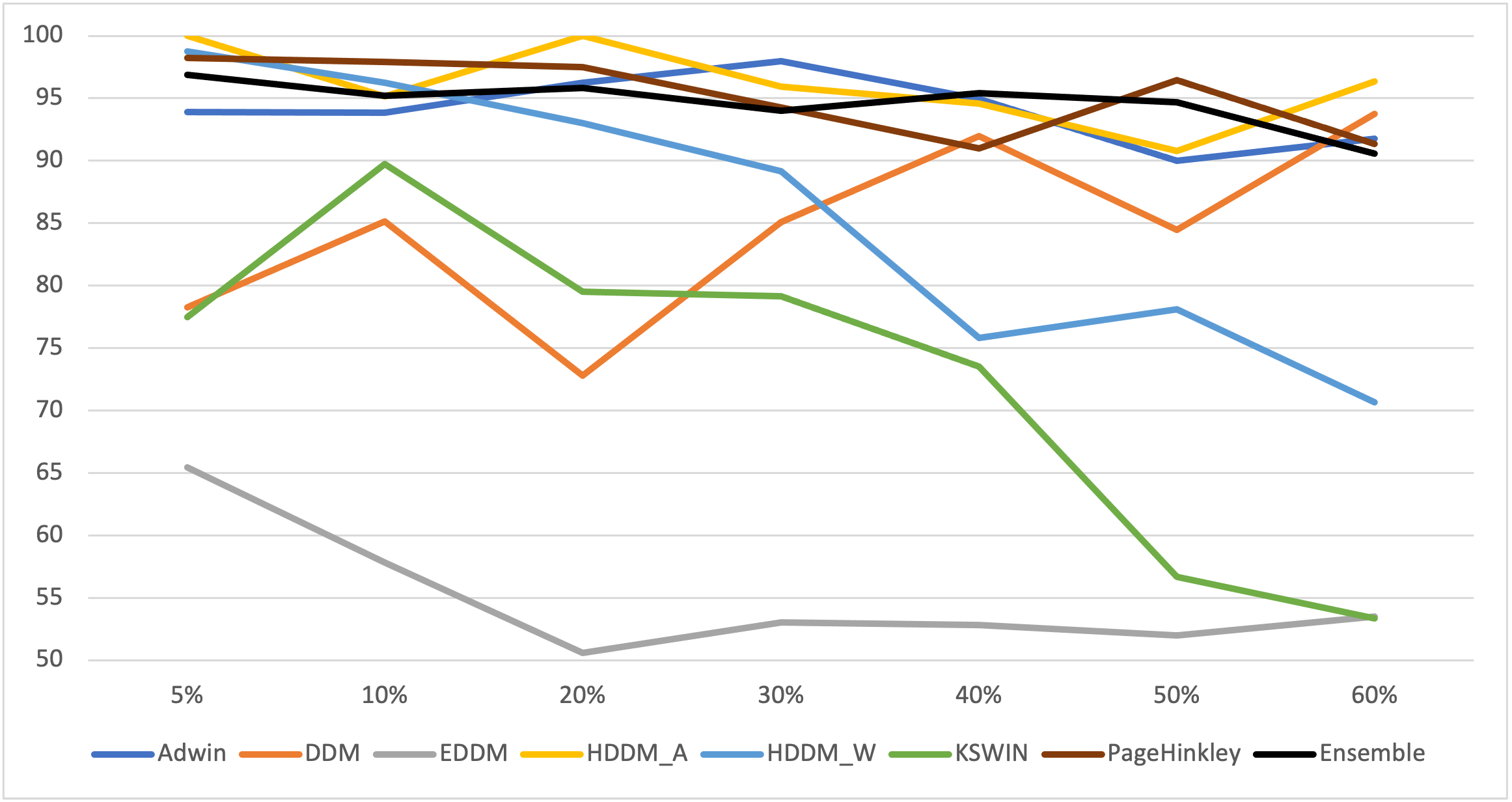}
  \subcaption{True positive rate.}
 \end{minipage}
 \newline
 \begin{minipage}[t]{.4\textwidth}
  \centering
  \includegraphics[width=\textwidth]{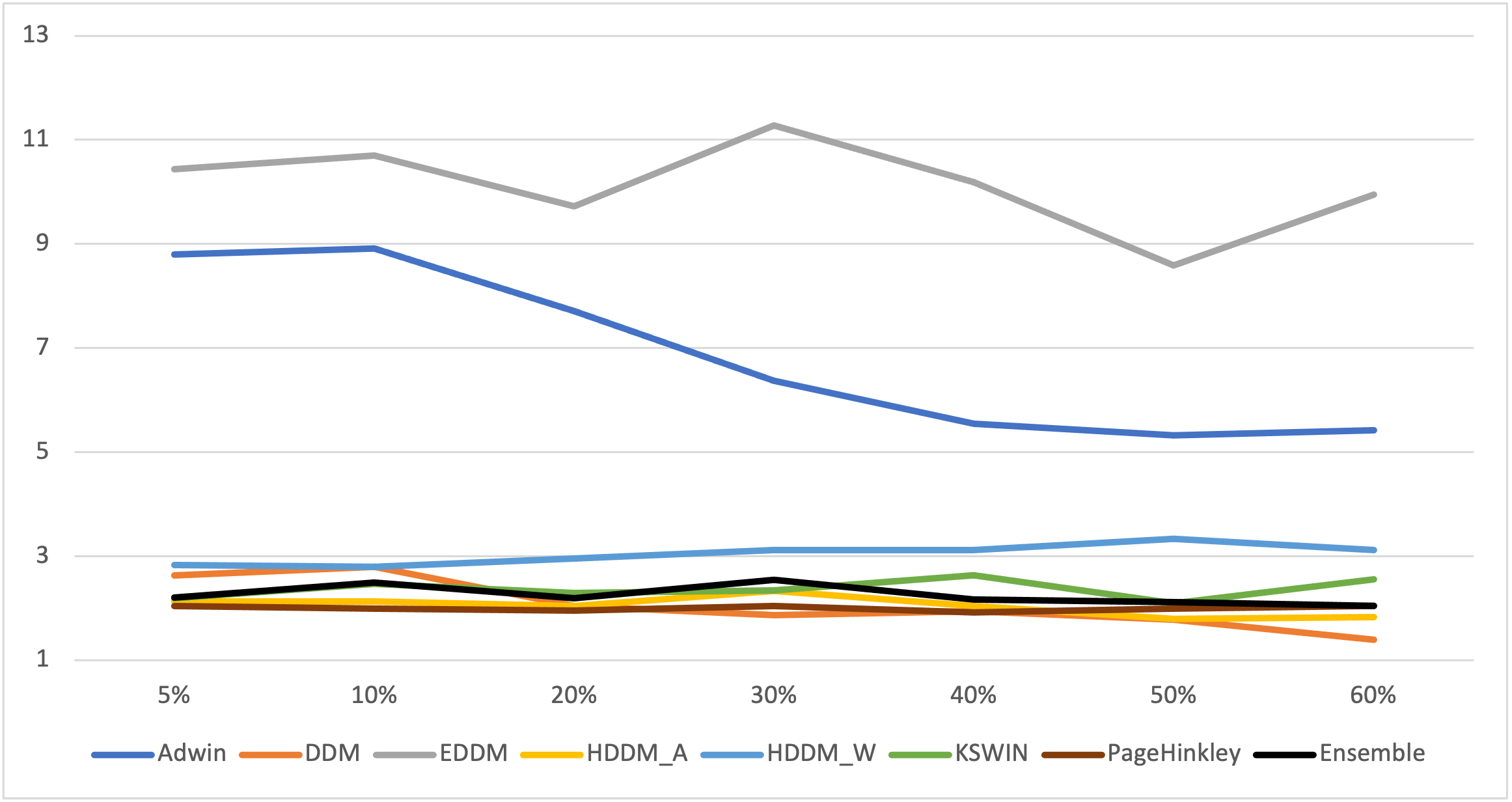}
  \subcaption{True positives per drift.}
 \end{minipage}
 \hfill
 \begin{minipage}[t]{.4\textwidth}
  \centering
  \includegraphics[width=\textwidth]{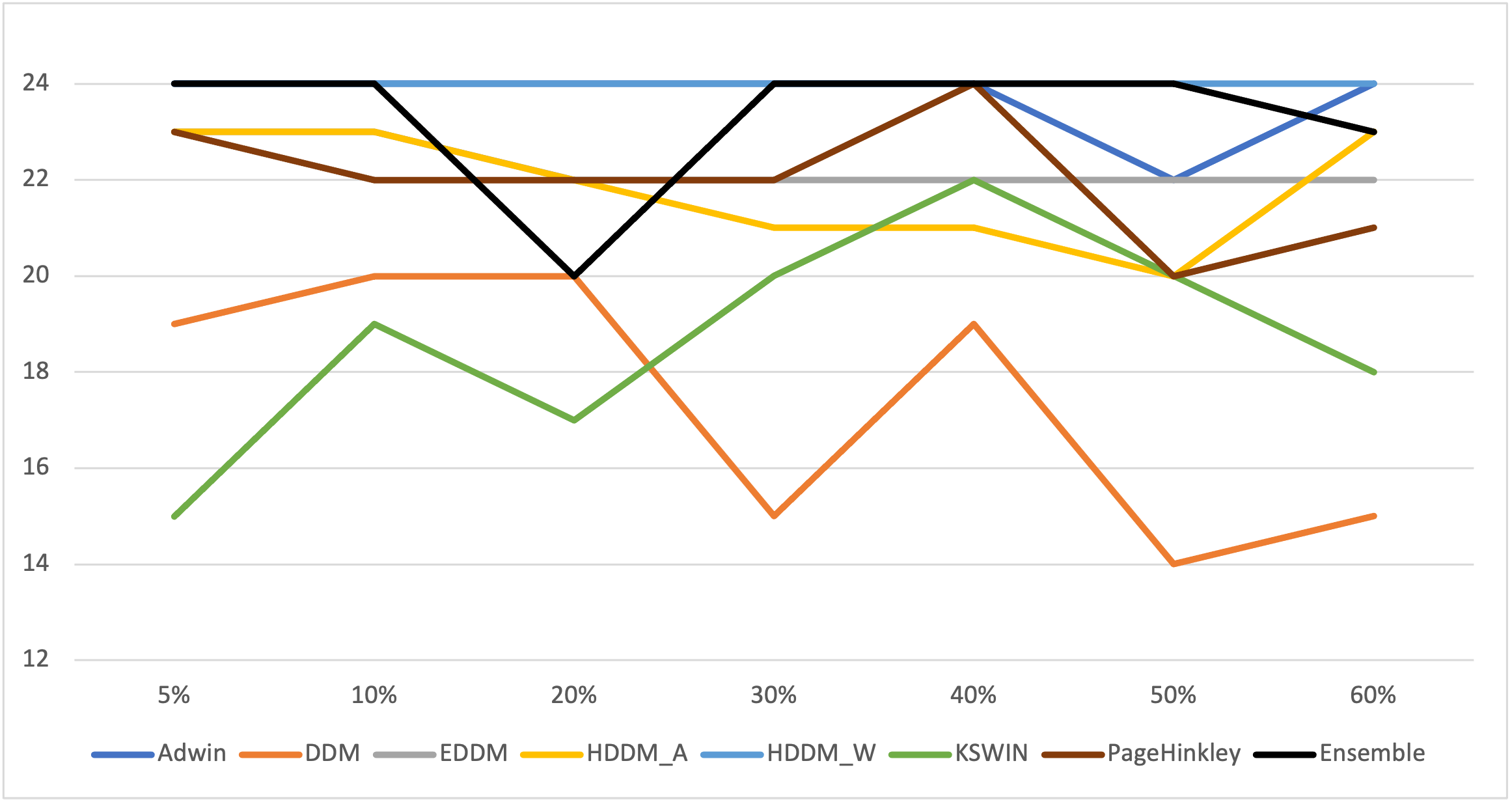}
  \subcaption{Drift count.}
 \end{minipage}
 \caption{Performance of CDDs (post data imputation) for gradual drifts in Harvard dataverse with different MAR sparsities.}
 \label{fig:sparsity}
\end{figure*}
Next we introduce different levels of sparsity using different techniques -- MCAR, MAR, MNAR -- on Harvard dataverse, and record the performance of the 7 concept drift detectors for the different metrics.
We notice that although prequential error is widely used as a metric for comparing CDDs~\cite{eddm,accurate,cdrift-review}; however, it can be sometimes misleading -- consider the case when one has a bad CDD that detects drifts often (even when there is none), and consequent retraining, which are typically quite costly, lead to low prequential error value.
Similar problem may arise with the metric \textit{accuracy}.
We have discussed the challenges with the other metrics already in Section~\ref{sec:metrics}. However, these metrics together serve as key indicators for performance of CDDs.

\begin{figure*}[tbh]
 \begin{minipage}[t]{.38\textwidth}
  \centering
  \includegraphics[width=\textwidth]{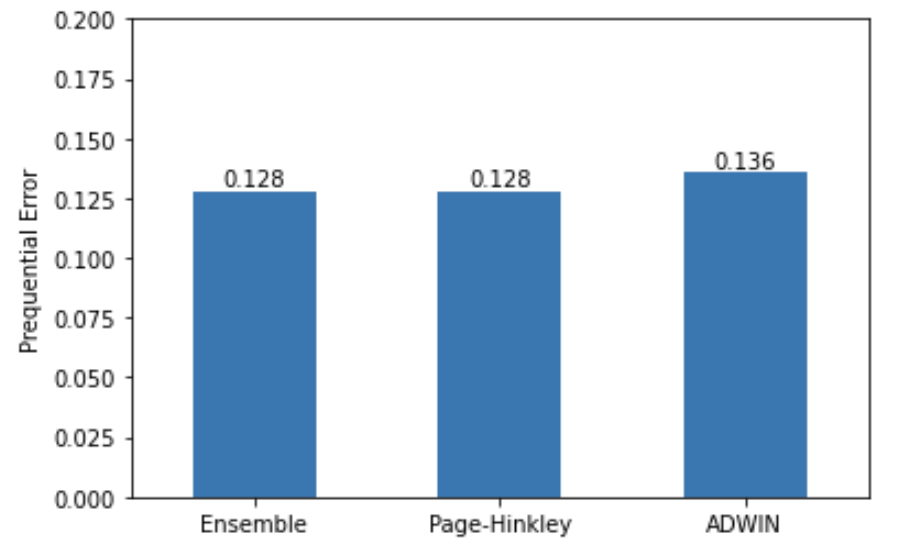}
  \subcaption{Prequential error.}
 \end{minipage}
 \hfill
 \begin{minipage}[t]{.38\textwidth}
  \centering
  \includegraphics[width=\textwidth]{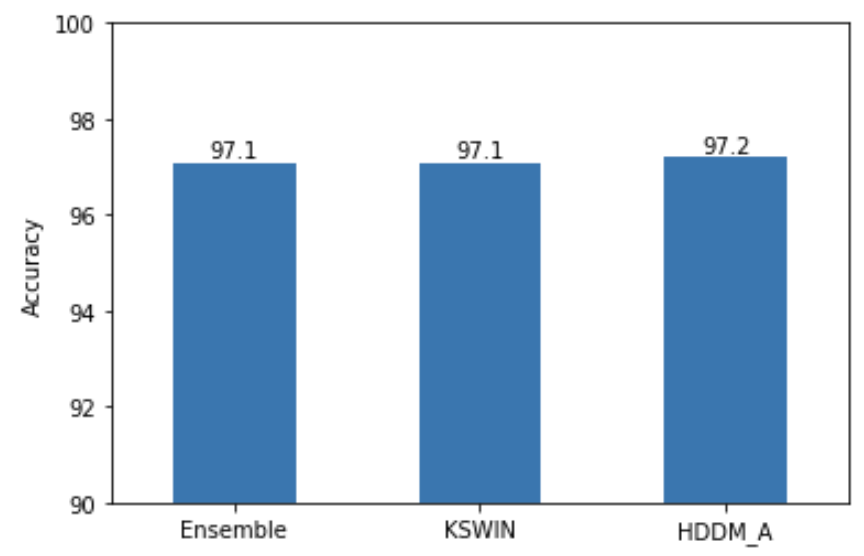}
  \subcaption{Accuracy.}
 \end{minipage}
 \newline
 \begin{minipage}[t]{.38\textwidth}
  \centering
  \includegraphics[width=\textwidth]{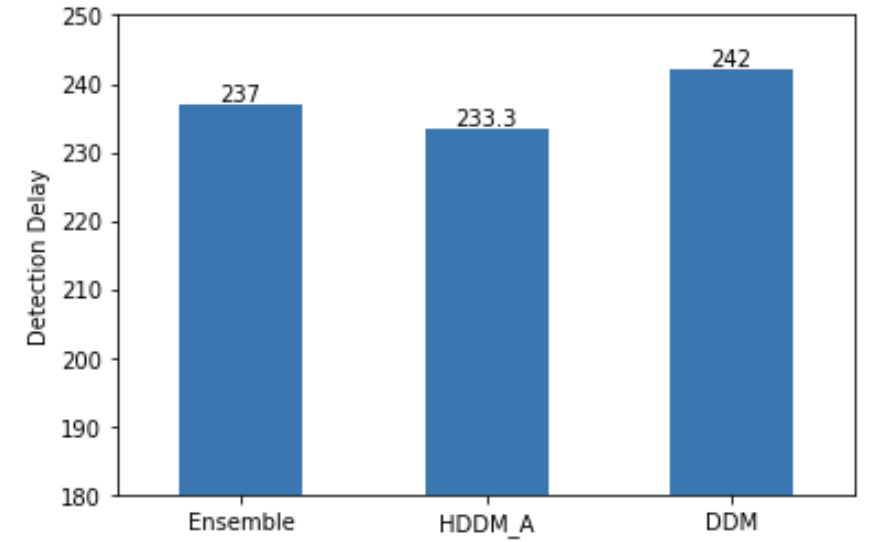}
  \subcaption{ADD.}
 \end{minipage}
 \hfill
 \begin{minipage}[t]{.38\textwidth}
  \centering
  \includegraphics[width=\textwidth]{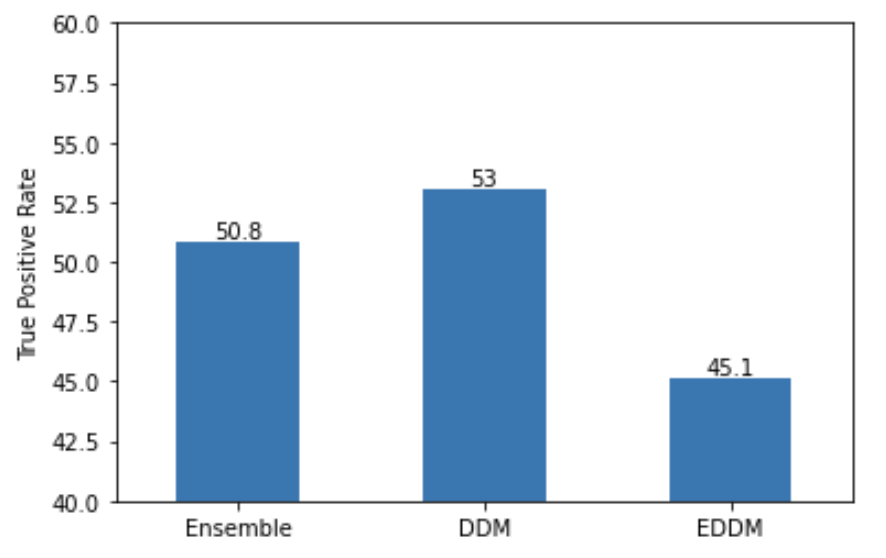}
  \subcaption{TPR.}
 \end{minipage}
 \newline
 \begin{minipage}[t]{.38\textwidth}
  \centering
  \includegraphics[width=\textwidth]{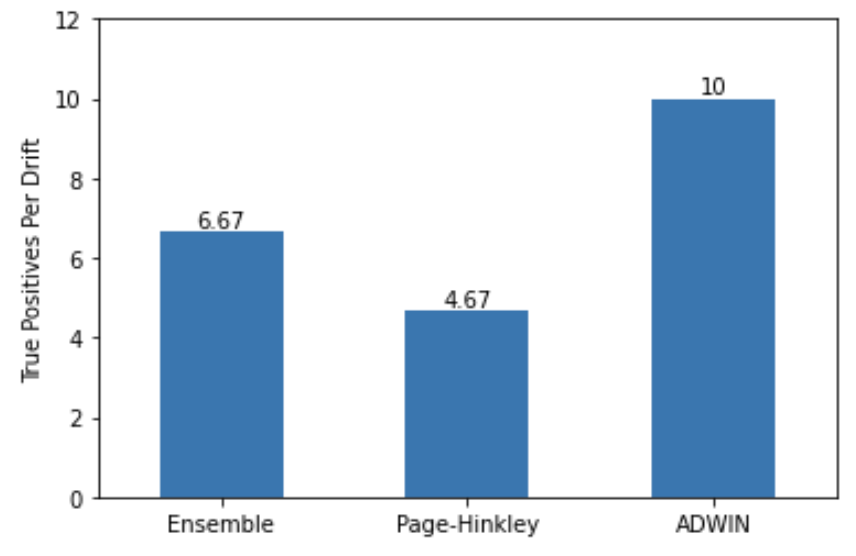}
  \subcaption{TPD.}
 \end{minipage}
 \hfill
 \begin{minipage}[t]{.38\textwidth}
  \centering
  \includegraphics[width=\textwidth]{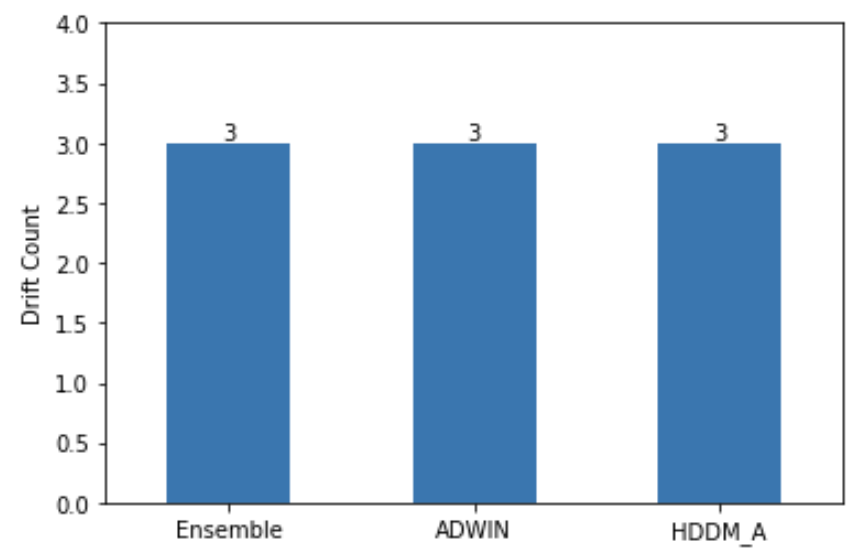}
  \subcaption{Drift count.}
 \end{minipage}
 \caption{Performance of top-3 CDDs (post data imputation) for gradual drifts in change risk assessment data.}
 \label{fig:changeRisk}
\end{figure*}

Empirically, we find that none of the 7 drift detectors performs well across all the metrics, as can be seen in Figure~\ref{fig:sparsity}, for any given missingness or drift type.
Hence, we decide to go with a majority voting based ensemble of detectors.
Ranking the algorithms based on their performance for different metrics, we find that for abrupt drifts: ADWIN $+$ HDDM\textsubscript{A} $+$ KSWIN, and for gradual drifts: HDDM\textsubscript{A} + HDDM\textsubscript{W} + Page-Hinkley, achieve best performance across all the metrics.
\par A challenge for working with an ensemble of detectors is finding the optimal window size, such that, if at any instant within that window, two of the detectors detect a drift, then the ensemble declares a drift at that instant.
Our experiments show that for Harvard dataverse, the optimal window size is 
$2000$, and for our data, it is $1000$.
The ensemble detector for Harvard dataverse is also shown in Figure~\ref{fig:sparsity}.
We omit the details of this experiment for finding the optimal window size due to page limitation.

We, finally, test our ensemble detector on change risk assessment data as shown in Figure~\ref{fig:changeRisk}.
As can be seen in this figure, if not the best, the ensemble detector always features in the top-3 for all the metrics.
Therefore, we plan to deploy it in near future because the majority voting based ensemble detector serves as a strong baseline, and it is unlikely that a concept drift will be missed by it.

\section{Conclusion}\label{sec:concl}
Concept drift detection poses a major challenge for real-world deployments of machine learning solutions.
The problem is further worsened in the presence of sparsity, especially since the ground truths for the missing values can never be obtained in reality.
One of our major findings in course of tackling this problem is that none of the popular concept drift detectors exhibit optimal performance across all metrics in all situations consistently.
Therefore, we design a majority voting based ensemble of detectors for abrupt and gradual drifts and it delivers best or close to best performance for the whole spectrum of metrics.

\bibliography{references}

\begin{thebibliography}{23}
\providecommand{\natexlab}[1]{#1}
\providecommand{\url}[1]{\texttt{#1}}
\providecommand{\urlprefix}{URL }
\expandafter\ifx\csname urlstyle\endcsname\relax
  \providecommand{\doi}[1]{doi:\discretionary{}{}{}#1}\else
  \providecommand{\doi}{doi:\discretionary{}{}{}\begingroup
  \urlstyle{rm}\Url}\fi

\bibitem[{Baena-Garcia et~al.(2006)Baena-Garcia, del Campo-Avila, Fidalgo,
  Bifet, Gavalda, and Morales-Bueno}]{eddm}
Baena-Garcia, M.; del Campo-Avila, J.; Fidalgo, R.; Bifet, A.; Gavalda, R.; and
  Morales-Bueno, R. 2006.
\newblock Early Drift Detection Method.
\newblock In \emph{Proc. 4th Int. Workshop Knowledge Discovery from Data
  Streams}.

\bibitem[{Bifet and Gavald{\`{a}}(2007)}]{adwin}
Bifet, A.; and Gavald{\`{a}}, R. 2007.
\newblock Learning from Time-Changing Data with Adaptive Windowing.
\newblock In \emph{{SDM}}, 443--448. {SIAM}.

\bibitem[{Bradley(1968)}]{runsTest}
Bradley, J.~V. 1968.
\newblock \emph{Distribution-Free Statistical Tests}.
\newblock Prentice-Hall.

\bibitem[{Breiman(2001)}]{Breiman}
Breiman, L. 2001.
\newblock Random Forests.
\newblock \emph{Mach. Learn.} 45(1): 5--32.

\bibitem[{Forbes et~al.(2010)Forbes, Evans, Hastings, and Peacock}]{Forbes}
Forbes, C.; Evans, M.; Hastings, N.; and Peacock, B. 2010.
\newblock \emph{Statistical distributions}.
\newblock John Wiley \& Sons, 4th edition.

\bibitem[{Frías-Blanco et~al.(2015)Frías-Blanco, Campo-Ávila,
  Ramos-Jiménez, Morales-Bueno, Ortiz-Díaz, and Caballero-Mota}]{hddm}
Frías-Blanco, I.; Campo-Ávila, J.~d.; Ramos-Jiménez, G.; Morales-Bueno, R.;
  Ortiz-Díaz, A.; and Caballero-Mota, Y. 2015.
\newblock Online and Non-Parametric Drift Detection Methods Based on
  Hoeffding’s Bounds.
\newblock \emph{IEEE Transactions on Knowledge and Data Engineering} 27(3):
  810--823.

\bibitem[{Gama et~al.(2004)Gama, Medas, Castillo, and Rodrigues}]{ddm}
Gama, J.; Medas, P.; Castillo, G.; and Rodrigues, P.~P. 2004.
\newblock Learning with Drift Detection.
\newblock In \emph{{SBIA}}, volume 3171 of \emph{LNCS}, 286--295. Springer.

\bibitem[{Gupta et~al.(2021)Gupta, Chatterjee, Matha, Banerjee, Parsai, and
  Agneeswaran}]{CD_Walmart}
Gupta, B.; Chatterjee, A.; Matha, H.; Banerjee, K.; Parsai, L.; and
  Agneeswaran, V. 2021.
\newblock Look Before You Leap! Designing a Human-Centered {AI} System for
  Change Risk Assessment.
\newblock \emph{CoRR} abs/2108.07951.

\bibitem[{Hastie et~al.(2015)Hastie, Mazumder, Lee, and Zadeh}]{softImpute}
Hastie, T.; Mazumder, R.; Lee, J.~D.; and Zadeh, R. 2015.
\newblock Matrix completion and low-rank {SVD} via fast alternating least
  squares.
\newblock \emph{J. Mach. Learn. Res.} 16: 3367--3402.

\bibitem[{Liu, Lu, and Zhang(2020)}]{cdrift}
Liu, A.; Lu, J.; and Zhang, G. 2020.
\newblock Concept Drift Detection: Dealing with MissingValues via Fuzzy
  Distance Estimations.
\newblock \emph{CoRR} abs/2008.03662.

\bibitem[{Lobo(2020)}]{harvardDataverse}
Lobo, J.~L. 2020.
\newblock {Synthetic datasets for concept drift detection purposes}.
\newblock \urlprefix\url{https://doi.org/10.7910/DVN/5OWRGB}.

\bibitem[{Lu et~al.(2019)Lu, Liu, Dong, Gu, Gama, and Zhang}]{cdrift-review}
Lu, J.; Liu, A.; Dong, F.; Gu, F.; Gama, J.; and Zhang, G. 2019.
\newblock Learning under Concept Drift: {A} Review.
\newblock \emph{{IEEE} Trans. Knowl. Data Eng.} 31(12): 2346--2363.

\bibitem[{Mitchell(1997)}]{Mitchell}
Mitchell, T.~M. 1997.
\newblock \emph{Machine Learning}.
\newblock McGraw-Hill.
\newblock ISBN 978-0-07-042807-2.

\bibitem[{Muzellec et~al.(2020)Muzellec, Josse, Boyer, and
  Cuturi}]{optimalTransport}
Muzellec, B.; Josse, J.; Boyer, C.; and Cuturi, M. 2020.
\newblock Missing Data Imputation using Optimal Transport.
\newblock In \emph{{ICML}}, volume 119 of \emph{Proceedings of Machine Learning
  Research}, 7130--7140.

\bibitem[{Page(1954)}]{pageHinkley}
Page, E.~S. 1954.
\newblock Continuous Inspection Schemes.
\newblock \emph{Biometrika} 41(1/2): 100--115.

\bibitem[{Paindaveine(2009)}]{multiRunsTest}
Paindaveine, D. 2009.
\newblock On Multivariate Runs Tests for Randomness.
\newblock \emph{Journal of the American Statistical Association} 104(488):
  1525--1538.

\bibitem[{Pesaranghader and Viktor(2016)}]{fastHoeffdig}
Pesaranghader, A.; and Viktor, H.~L. 2016.
\newblock Fast Hoeffding Drift Detection Method for Evolving Data Streams.
\newblock In \emph{Machine Learning and Knowledge Discovery in Databases},
  96--111.

\bibitem[{Raab, Heusinger, and Schleif(2020)}]{kswin}
Raab, C.; Heusinger, M.; and Schleif, F. 2020.
\newblock Reactive Soft Prototype Computing for Concept Drift Streams.
\newblock \emph{Neurocomputing} 416: 340--351.

\bibitem[{Rubin(1976)}]{Rubin}
Rubin, D.~B. 1976.
\newblock {Inference and missing data}.
\newblock \emph{Biometrika} 63(3): 581--592.

\bibitem[{Sethi and Kantardzic(2017)}]{reliableDetection}
Sethi, T.~S.; and Kantardzic, M.~M. 2017.
\newblock On the reliable detection of concept drift from streaming unlabeled
  data.
\newblock \emph{Expert Syst. Appl.} 82: 77--99.

\bibitem[{Souza et~al.(2020)Souza, Reis, Maletzke, and
  Batista}]{challengesRealData}
Souza, V. M.~A.; Reis, D.~M.; Maletzke, A.~G.; and Batista, G. E. A. P.~A.
  2020.
\newblock Challenges in Benchmarking Stream Learning Algorithms with Real-world
  Data.
\newblock \emph{Data Mining and Knowledge Discovery} 34: 1805--1858.

\bibitem[{van Buuren and Groothuis-Oudshoorn(2011)}]{iterativeImputer}
van Buuren, S.; and Groothuis-Oudshoorn, K. 2011.
\newblock mice: Multivariate Imputation by Chained Equations in R.
\newblock \emph{Journal of Statistical Software, Articles} 45(3): 1--67.
\newblock ISSN 1548-7660.

\bibitem[{Yan(2020)}]{accurate}
Yan, M. M.~W. 2020.
\newblock Accurate detecting concept drift in evolving data streams.
\newblock \emph{ICT Express} 6(4): 332--338.

\end{thebibliography}

\end{document}